\documentclass[10pt,twocolumn,letterpaper]{article}
\usepackage{cvpr}              
\definecolor{cvprblue}{rgb}{0.21,0.49,0.74}
\usepackage[pagebackref,breaklinks,colorlinks,allcolors=cvprblue]{hyperref}

\def\paperID{*****} 
\def\confName{CVPR}
\def\confYear{2026}

\title{Realtime-VLA V2: Learning to Run VLAs Fast, Smooth, and Accurate}

\author{Chen Yang \and Yucheng Hu \and Yunchao Ma \and Yunhuan Yang \and Jing Tan \and Haoqiang Fan \\
Dexmal\\
{\tt\small \{yangchen, hyc, myc, yyh, tanjing, fhq\}@dexmal.com}
}

\begin{document}
\maketitle
\begin{abstract}
In deployment of the VLA models to real-world robotic tasks, execution speed matters. In previous work~\cite{ma2025running} we analyze how to make neural computation of VLAs on GPU fast. However, we leave the question of how to actually deploy the VLA system on the real robots open. In this report we describe a set of practical techniques to achieve the end-to-end result of running a VLA-driven robot at an impressive speed in real world tasks that require both accuracy and dexterity.
The stack of technology ranges across calibration, planning \& control, and learning based method to identify optimal execution speed. In the tasks we show, the robot even executes in a speed on par with casual human operation and approaching the hardware limit of our lightweight arm. The unaccelerated videos and inference traces are provided in \url{https://dexmal.github.io/realtime-vla-v2/}.
\end{abstract}    
\section{Introduction}
\label{sec:intro}

In real-world robotic tasks, cycle time matters a lot. The effort to make VLA-based robot execution fast spans across asynchronous control methods, adaptive velocity planning, optimization of training data, and high performance computing. In this report, we try to put the ingredients together to see how fast we can actually make a robot move with current VLAs.

The first obstacle resides in the robot hardware. Typical robotic arms are capable of high maximum joint speeds, but they do not endure abrupt acceleration or jerky motion. When we speed-up the action chunk, the hardware system itself begins to be unstable, especially for the popular QDD-based light-weight arms that are usually low in stiffness.

The next conceptual difficulty comes from the imitation learning principle. When the robot records demonstration data with human operated tele-operation, its speed is drastically reduced. Recording high-speed demonstration trajectories is hard for human due to hardware limit and delayed control feedback. So we must step in the realm of faster-than-demonstration execution of the learned policies, and this creates mismatch in environment dynamics between training and inference.

Finally, we have to achieve speed-up without compromising one-time task success rates. If the robot needs to frequently retry an action or even the whole task, the time spent in failure recovery will offset the performance gain in terms of end-to-end task throughput.

The observations above give us a more precise characterization of the ``making robot fast'' problem: We want an approach that runs the robot \textbf{fast, smooth and accurate}. Obtaining smooth trajectories is the prerequisite of hardware-friendly high velocity execution. Keeping the trajectories accurate is crucial for task success. Achieving these three goals at the same time is the key for a practical faster-than-demonstration system.

Given the analysis, we solve this problem in a systematic approach. We combine the well-established components in robotics to attack the issues one-by-one, and construct a set of techniques that work end-to-end in practice. For the most critical part, we resort to learning-based method and use a model.

The effect of our strategies is manifested in our demo videos. The robots can move in a rapid and fluent fashion, free of shakes, and the end-to-end execution speed is even approaching human's operation (with the constraint that the human must also work in the way reminiscent to the robots, e.g. using grabs instead of in-hand manipulation). We also observe that the techniques are generalizable to a broad range of tasks, even including task that requires very high precision of motion.

Our contributions are as follows:
\begin{itemize}
    \item We identify the core issues in faster-than-demonstration execution of VLA policies, and describe an implementation framework to handle them;
    \item We explore how experience data can be utilized for learning-based optimization of task throughput;
    \item We give an analysis of the upper-bound on VLA action execution speed.
\end{itemize}

\begin{figure*}
    \centering
    \includegraphics[width=1.0\linewidth]{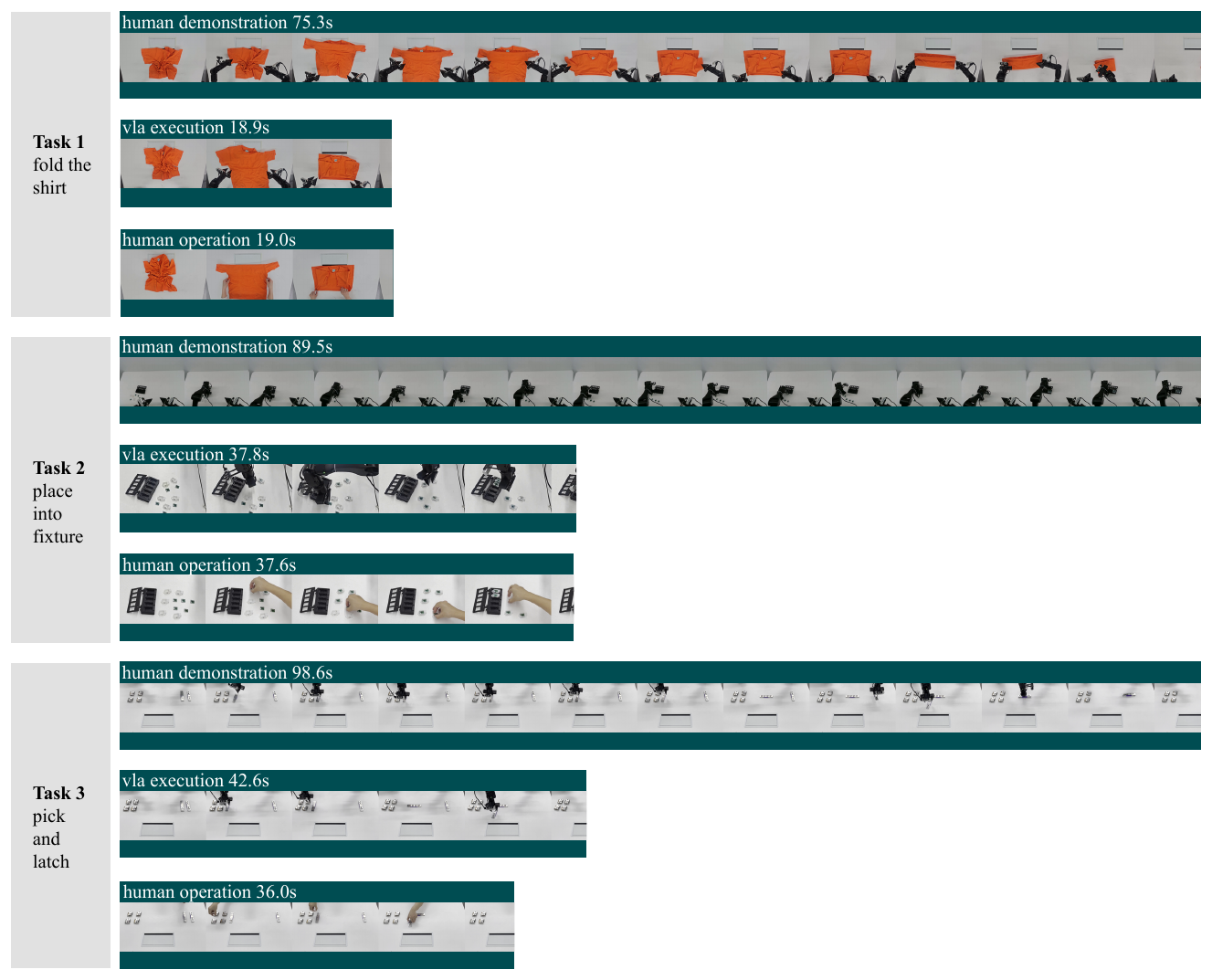}
    \caption{Example tasks used in this paper. See \url{https://dexmal.github.io/realtime-vla-v2/} for unaccelerated videos.\label{fig:speedcompare}}
\end{figure*}

For the organization of the following sections, we will cover the ingredients in our system in a step-by-step manner, and then give the details of our experiments. We will also demonstrate how to construct a ``upper bound'' analysis for robot execution speed.


\section{Related Works}
\label{sec:relatedwork}
\subsection{Vision-Language-Action Models}
Vision-language-action (VLA) models provide a powerful paradigm for robotic manipulation, mapping visual observations and language instructions to low-level actions. Representative works include RT-1 \cite{brohan2022rt}, RT-2 \cite{zitkovich2023rt}, PaLM-E \cite{driess2023palm}, ACT \cite{zhao2023act}, Octo \cite{team2024octo}, OpenVLA \cite{kim2024openvla}, and Gr00t \cite{bjorck2025gr00t}, which enable strong cross-task and cross-embodiment generalization.

Despite advances in semantic reasoning, these systems typically inherit the slow temporal characteristics of human demonstrations, resulting in practical cycle-time bottlenecks. Recent works such as $\pi$ \cite{black2025trainingrtc} and VLASH \cite{tang2025vlash} improve efficiency via high-frequency action generation and asynchronous inference.

However, these methods focus on policy generation rather than system-level constraints including control latency, trajectory smoothness, and hardware dynamics. In contrast, we accelerate VLA execution from a system-level perspective, integrating asynchronous control, trajectory shaping, and speed adaptation to enable reliable faster-than-demonstration operation.

\subsection{Imitation Learning and Execution Acceleration}
Imitation learning is constrained by the speed and distribution of human demonstrations. Prior works such as DAgger \cite{ross2011dagger} and ACT \cite{zhao2023act} improve generalization and stability via dataset aggregation and stronger policy representations.

Recent research focuses on faster-than-demonstration execution. Data-level methods including ESPADA \cite{kim2025espada} and DemoSpeedup \cite{guo2025demospeedup} adjust demonstration trajectories to encourage faster motion patterns. Policy and speed adaptation methods such as SpeedAug \cite{nam2025speedaug}, SAIL \cite{arachchige2025sail}, and RLT \cite{xu2025rltoken} jointly optimize speed and stability via data augmentation and online fine-tuning.

In contrast, our method uses human-modulated rollouts to learn fine-grained speed adaptation, either implicitly within the VLA policy or via a lightweight adapter, enabling flexible and plug-and-play speed modulation without architectural changes.

\subsection{System-Level Execution and Control}
Classical trajectory optimization methods including CHOMP \cite{ratliff2009chomp}, STOMP \cite{kalakrishnan2011stomp}, TrajOpt \cite{schulman2014motion}, and MPC \cite{garcia1989model} generate smooth and dynamically feasible motions. Recent works \cite{oelerich2026safeflowmpc} integrate learning-based policies as reference trajectories for tracking and refinement.

However, these approaches focus on accurate execution rather than reliable high-speed operation. When accelerating beyond demonstration speed, system bottlenecks including control latency, sensing–actuation delay, and trajectory smoothness dominate stability, yet remain largely unaddressed.

In contrast, we identify temporal alignment, trajectory shaping, and closed-loop control as critical for reliable high-speed VLA execution, and jointly optimize these components to enable stable, smooth, and accurate operation beyond demonstration speed in real-world settings.


\begin{figure}
    \centering
    \includegraphics[width=0.99\linewidth]{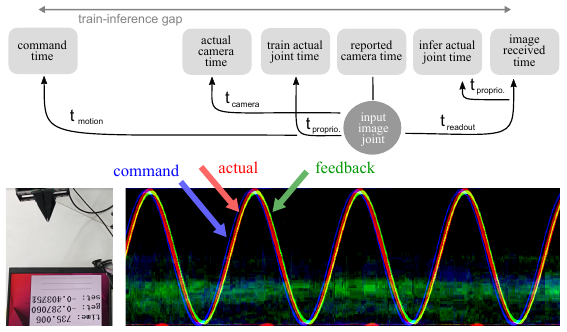}
    \caption{Time delay in the system and train-inference discrepancy.\label{fig:delaygraph}}
\end{figure}

\section{Calibration and Identification of the System}
\label{sec:system}

\subsection{Delays in the System}

The first thing we need to get right is to understand the time delay in the system. In VLA training, we present the model an ideal environment where observation is instant and actions take place immediately. From a training perspective, this is desired: The model will be confused by unaligned signals and generate blurred trajectories otherwise. However, in real world there are a handful of temporal misalignment on stage.

When we record training data, the camera gives a timestamp for each frame (the reported camera time), which will be used to align with joint position readings to form training pairs. However, we find the actual exposure can happen $t_\text{camera}$ before the timestamp is taken, and the joint reading may be delayed by $t_\text{proprio.}$. Also, the image will be received $t_\text{readout}$ after the claimed timestamp. All these time delays can be reduced to minimum amount if all components are well-engineered. But for off-the-shelf cameras and robots, they may take time on the magnitude of tens of milliseconds.

Another more significant delay comes from the dynamics of the robot. When we control it in Cyclic Synchronous Position (CSP) mode, we send the robot a target position at each tick. However, the actual position of the robot will lag behind. The lag is partly due to communication delay, but the biggest contribution comes from the tracking algorithm of the arm. We observe that many lightweight arm uses PD control (or even P control) in the position loop for maximum stability, so there will be a lag inverse proportionally to the gain of the PD controller. These creates a constant-time delay for linear motions (we call it $t_\text{motion}$), and an exponential decay for stops or starts.

The relationship between the terms above is summarized in \cref{fig:delaygraph}. We can see that compared to an idealistic model of ``generating next moment's action when seeing the image'', there is a total of 
 $t_\text{motion} + t_\text{proprio.} + t_\text{readout}$
gap in the real world. This discrepancy explains many of the nuisances we observe in the experiments if we do not account for them.

\subsection{Calibration and Compensation}

Given the issues above, we now need to devise a way to measure the time delays so we can handle them later.

The easiest is $t_\text{readout}$, which can be obtained just by comparing system and camera timestamps. For $t_\text{camera}$, we setup a strip of LEDs driven by system clock and compare captured content with the claimed timestamp to measure latency. These are the well established methods in camera calibration.

For the robot-related delays, measurement is harder when we are not equipped with high precision measurement devices. However, we find a practical way to resolve them to good accuracy. The setup is shown in the lower left part of \cref{fig:delaygraph}. We let the robot sway in a sinusoidal pattern in front of the camera. Then we display on a screen three track bars representing the phase (with respect to the sin cycle) of the current time, the actual position as reported by the robot, and the commanded target position as we send to the robot. The setup is recorded by a high fps camera (many mobile phones supports up to 120fps), and we extract from each frame the line of image of the track bar and arm tip area to form a time-position plot. An example is shown in the lower right part of \cref{fig:delaygraph}, in which we use the three image channels to represent the three measured movement.

The advantage of using imagery data is that we can do phase estimation from the pixel values to achieve sub-frame accuracy. From our experiments, we can calibrate $t_\text{motion}$, $t_\text{proprio.}$ up to 5 milliseconds. Note that we have to account for screen display delay. This can be done by comparing the latency measured by the system time track bar (which adds screen latency to camera latency) and the latency by the LED measurements (which only has camera latency).

We show the concrete figures we obtained with our robot setup, the DOS W1 system, which uses RealSense D435 camera and Airbot Play arm.

\begin{center}
\begin{tabular}{c|c}
delay & measurement \\
\hline
$t_\text{readout}$ & 33ms \\
$t_\text{camera}$ & 55ms \\
$t_\text{proprio.}$ & 50ms \\
$t_\text{motion}$ & 150ms \\
\end{tabular}
\end{center}

After measuring the latencies, we now describe how to compensate for them. For $t_\text{camera}$ and $t_\text{readout}$, we combine them into a single latency standing for the time that has passed when we receive the image. Then we use this value to retrieve from a small history buffer of the reported joint positions (taking into account $t_\text{proprio.}$.). In this way we make sure that the image and joint readings are aligned for the input to the VLA.

As for $t_\text{motion}$, sending the VLA commands ``ahead of time'' does not help, as it is essentially the result of a control loop instead of a constant transmission delay. We put this parameter in the post-processing stage of the model trajectory (detail in the next section) to pre-amplify the target positions so that the robot will closely track the model's output.

\begin{figure}
\centering
\includegraphics[width=0.99\linewidth]{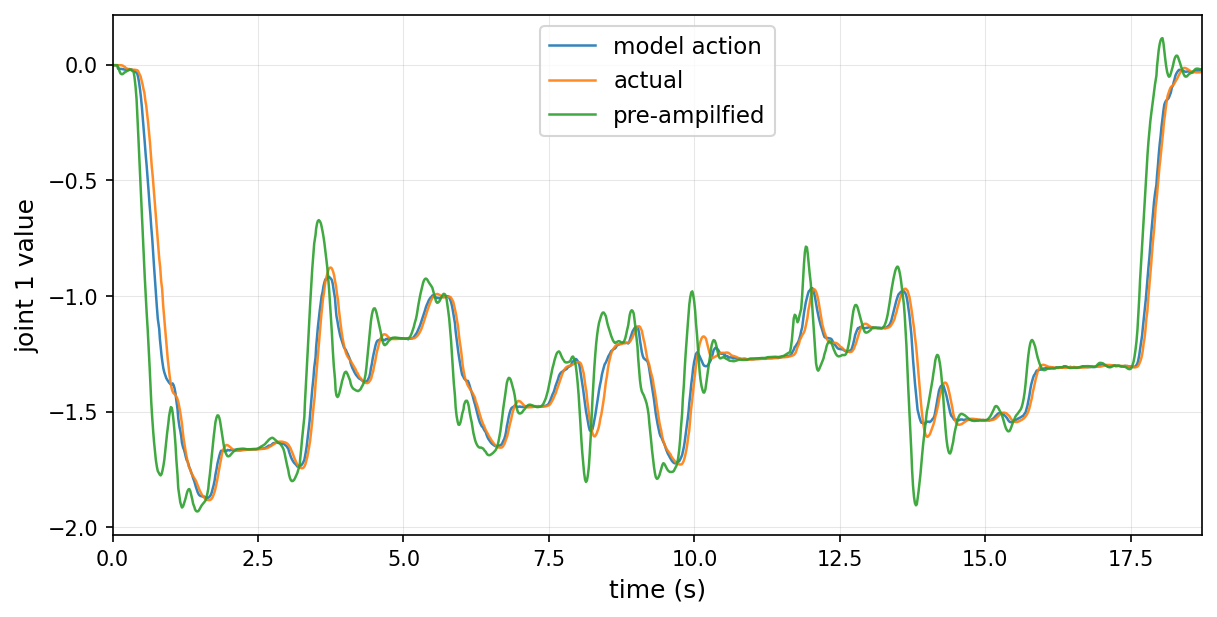}
\caption{Effect of pre-amplification of robot command. We have to increase the magnitude of change in the robot command so that the actual position tracks model's target. \label{fig:preamplify_vis}}
\end{figure}

In \cref{fig:preamplify_vis} we show the effect of pre-amplification. The target position command we send to the robot is way more drastic than the model's desired path. After the robot's controller, the trajectory comes back to what the model really demands. This is the key step to align model's view of the timeline to the lagging real world.

\section{Post-processing the VLA Trajectory}
\label{sec:postprocessing}
Our core idea is to post-process the VLA's output trajectory in a way that is transparent to the model. We want to give the model a view as if its actions are taken place perfectly so that we do not need to re-train the model and the asynchronous techniques like RTC are still valid. This can be done by \textbf{temporally} adjusting the trajectory -- re-assigning the durations to the time steps so that the trajectory becomes smooth, and then speed up the smoothed trajectory for maximum speed.

\begin{figure}[t]
    \centering
    \includegraphics[width=0.99\linewidth]{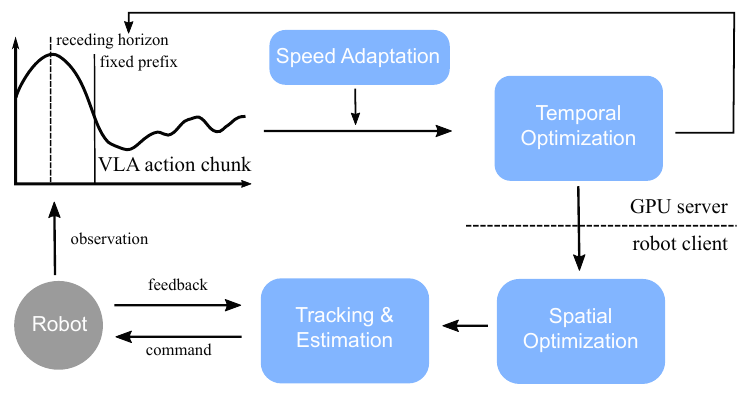}
    \caption{Post-processing framework of VLA's trajectory.}
    \label{fig:postprocessing_framework}
\end{figure}

We want to emphasize that \textbf{smoothness} of the trajectory is the most critical aspect in high speed execution. The real-world obstacle of running fast is not the maximum speed of the joint -- it is that the system should not diverge when running in high speed. The jitters and vibrations in the control signal will make the input visual shaky and destabilize both the VLA model and the joint hardware. However, if we can make the trajectory smooth, the hardware capacity can be utilized to its maximum while it tightly track the target positions in the trajectory.

Therefore, we implemented the trajectory post-processing framework shown in \cref{fig:postprocessing_framework}. We assumed a client-server separation where the VLA model is run on another GPU machine (the server) and the robot is connected to a local computer or board (the client). The VLA's trajectory is first modulated by a Speed Adaptation Model, then it is processed by a Temporal Optimization model and sent to the client. Afterwards, the client does Spatial Optimization at each tick of execution, and then tracks and estimates the state of the robot through control and feedback. The detail of each step will be described below.

\subsection{Speed Adaptation and Temporal Optimization}

The first phase operates on the \textbf{temporal} profile of the trajectory, without changing the shape of the path that VLA generates. We put this restriction because (1) We don't want to lose the positional accuracy of the way-points that VLA gives (2) this works independent of the prefix-matching mechanism of real-time chunking. From the perspective of the VLA model, the temporal adjustments merely changes how fast the action chunk is consumed, so it will not introduce inconsistency to the visual control loop.

In this phase, a \textbf{Speed Adaptation Model} is used to decide the per-step speed of the chunk (in the form of a scaling factor). We will be describing how such a model can be trained in latter sections. The model is supposed to dynamically accelerate the execution to the limit that success rate is preserved.

However, we cannot rely on a model to guarantee smoothness. In practice we find that the path contains short segments that some joint rotates abruptly, creating a high acceleration peak. We need to distribute the acceleration throughout the chunk, without altering the average speed. We do this in an optimization framework, so we call it \textbf{Temporal Optimization}. Specifically, we formulate a quadratic program over the inverse of step duration $\bigtriangleup t_{i}^{-1}$:

\begin{align}
\min_{\bigtriangleup t_{1:H}} \quad & \lambda_{0} \sum_{i=1}^{H} \| \bigtriangleup t_i^{-1} - (\bigtriangleup t_i^{ref})^{-1}\|^2 + \lambda_{1} \|a_{i}\|^2 \\
\text{s.t.} \quad & \bigtriangleup t_{\min} \leq \bigtriangleup t_i \leq \bigtriangleup t_{\max} \\
                  & v_i \leq v_{\max}
\end{align}

where $v_{i}$ and $a_{i}$ are defined as
\begin{align}
v_{i} & = \frac{(s_{i+1} - s_{i})}{\bigtriangleup t_{i}} \\
a_{i} & = \frac{(s_{i+1}  + s_{i-1} - 2*s_{i})}{\bigtriangleup t_{i}}
\end{align}
$s_{i}$ denotes the waypoints generated by the VLA model, $\bigtriangleup t_i^{ref}$ is obtained by converting the velocity scaling factor predicted by the speed adaptation model. $\lambda_{0}$, $\lambda_{1}$, $\bigtriangleup t_{\min}$, $\bigtriangleup t_{\max}$, $v_{\max}$ are all configurable parameters. We solve this problem by OSQP.

In this way, we should obtain a chunk at an appropriate speed and is free of abrupt acceleration. This trajectory is then sent to the client side through network.

\subsection{Spatial Optimization and Tracking}

At this stage, we will diverge from the temporal-only paradigm and alter the positions of the trajectory locally. The goal of this step is two fold: (1) Ensuring the trajectory is valid w.r.t. actual hardware constraint (2) Driving the robot to track the trajectory. These two goals are implemented in one single process, which we called \textbf{Spatial Optimization}, and it is essentially a control algorithm.

We model the lagging dynamics of the robot system by a linear recurrent relationship:
\[
\mathbf q_{k+1} = a \mathbf q_k + (1-a)\mathbf y_k,
\qquad
a = \exp(-h/\tau),
\]
where \(h\) is the servo period and \(\tau\) is the identified response time constant. This model captures the dominant effect that matters for high-speed execution: the realized motion lags behind the issued command and attenuates abrupt command changes.

We can use this relationship to compensate for the delay in joint position observation: We record a history of commands after the actual timestamp of the last received state feedback, and replay the effects of the commands through the equation above. Then we have a corrected observation of the current state.

Finally we will decide the current command to send, given the current state $q_{0}$, previous commands$y_{-1}, y_{-2}$, and the future targets $r_k$:
\[
\begin{aligned}
\min_{\{\mathbf y_k\}_{k=0}^{N-1}} \quad
&
\sum_{k=0}^{N-1}
\Big[
\|\mathbf q_k-\mathbf r_k\|_2^2
+
\lambda_{\mathrm{cmd}}\|\mathbf y_k-\mathbf r_k\|_2^2
\\
&\hspace{2em}
+
\lambda_{\mathrm{lag}}\|\mathbf y_k-\mathbf q_k\|_2^2
+
\lambda_{\Delta}\|\mathbf y_k-\mathbf y_{k-1}\|_2^2
\\
&\hspace{2em}
+
\lambda_{\Delta^2}\|\mathbf y_k-2\mathbf y_{k-1}+\mathbf y_{k-2}\|_2^2
\Big]
\\
&\hspace{1em}
+
\lambda_f \|\mathbf q_N-\mathbf r_N\|_2^2
\\
\text{s.t.}\quad
&
\mathbf q_{k+1}=a\mathbf q_k+(1-a)\mathbf y_k,
\\
&
\mathbf q_{\min}\le \mathbf y_k\le \mathbf q_{\max},
\\
&
-\mathbf d_{\max}\le \mathbf y_k-\mathbf q_k\le \mathbf d_{\max},
\\
&
-h\mathbf v_{\max}\le \mathbf y_k-\mathbf y_{k-1}\le h\mathbf v_{\max},
\\
&
-h^2\mathbf a_{\max}\le
\mathbf y_k-2\mathbf y_{k-1}+\mathbf y_{k-2}
\le h^2\mathbf a_{\max},
\\
&
k=0,\dots,N-1 .
\end{aligned}
\]

The objective has a direct physical interpretation. The term \(\|\mathbf q_k-\mathbf r_k\|_2^2\) makes the optimizer track the desired realized trajectory rather than the raw command sequence. The auxiliary command-tracking term \(\|\mathbf y_k-\mathbf r_k\|_2^2\) keeps the solution close to the reference trajectory from the cloud side when no compensation is needed, while \(\|\mathbf y_k-\mathbf q_k\|_2^2\) and the bound \(\mathbf d_{\max}\) prevent overly aggressive pre-amplification. The first- and second-difference penalties regularize command velocity and curvature, suppressing sharp command jumps, jitter, and high-frequency oscillations that become particularly harmful at elevated execution speeds.

We implement this module using an \texttt{acados} \cite{verschueren2022acados} MPC planner in SQP-RTI mode. This choice enables low-latency, repeated re-planning in the inner control loop.

We stress that the Spatial Optimization is needed even though we have already smoothed the trajectory in the Temporal Optimization part. Firstly, the temporal optimization does not apply hard constraint and is mainly driven by a model, so we cannot solely rely on it. Secondly, the actual real-time joint states are only taken into account at this part, so it needs to be running on the client side. Theoretically, if the trajectory is already smooth enough, and the robot follows the command perfectly, this optimization phase will be a no-op. However, when we push the execution speed of the robot to its limit this step becomes mandatory.
\section{Learning When and How to Speedup}
\label{sec:speedup}

\begin{figure}
    \centering
    \includegraphics[width=0.99\linewidth]{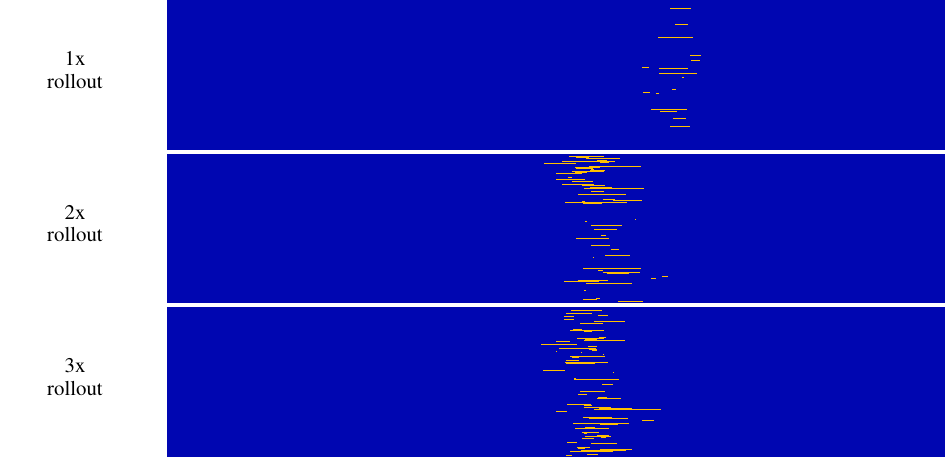}
    \caption{Distribution of failure timestamps during autonomous rollouts at $1\times$, $2\times$, and $3\times$ execution speeds. At each speed, failures cluster around a distinct bottleneck stage, revealing a small set of phases that dominate the overall failure rate.}
    \label{fig:failure_distribution_speed}
\end{figure}

In the Speed Adaptation module, we need to adjust the velocity of the actions based on what it is doing. Empirically, we observe that most part of the execution trajectory tolerates a high speed-up factor. It can goes up to 3x or even 4x depending on the task. However, there are critical stages in the task that requires slow-down. To get a better understanding of this issue, we show an experiment in Fig.~\ref{fig:failure_distribution_speed}. In this experiment, after training a model for the fold-the-shirt task, we run it with different speed-up factors (1x, 2x and 3x). When it runs into failure, we let a human operator to intervene the execution, and plot these regions in the figure. As can be seen, when running at a high speed-up factor, the number of interventions increased dramatically, yet they concentrate on a very specific stage of the whole task. For the fold-the-shirt task, we observe the most sensitive actions are folding the two sleeves, as they require high precision and a very specific velocity to avoid slipping.

This motivates us that we should have an ``intelligent'' module to dynamically adjust the execution speed, and the knowledge of deciding the optimal speed factor can be learned from \textbf{experience}~\cite{silver2025welcome}. This is different from previous work where geometric heuristics or grabber states are used alone. Also, this will be more precise than coarsely segmenting the run into ``casual'' or ``precise'' parts with common sense and hand-tuning the speed-up factors.

In the next subsections, we describe different possible solutions to learning the model.
\subsection{Failure-rate Prediction Model}

An obvious method is to learn a Q function to the speed modulation problem: Given a set of observation and a set of speed-up factor candidates, we learn the probability that the execution will fail if it is run at this particular speedup. Then at inference time we select the best speed-up factor as the maximum one which leads to a tolerable failure probability.

This method is intuitive, and a good thing is that it can be learned from rollout-only data. Using the trained model and a predefined uniform speed-up factor, the data collection operator only needs to overlook the execution process and stop the robot when it fails. This puts minimum requirement on the human operator, so this kind of data can be collected reliably.

However, in practice we still find difficulties in the implementation of this idea. The model is prone to overfitting as the rollout data is inevitable scarce. The distribution of input observations are biased as in high speeds some states will be masked out by previous failures so that are never visited. Tweaking the threshold to determine the best factor from failure probability vector is tricky.

All these nuisances above lead us to propose a new strategy for learning.

\subsection{Human-in-the-loop Speed Modulation Data Collection}
\label{sec:data_collection}
\begin{figure}
    \centering
    \includegraphics[width=1.0\linewidth]{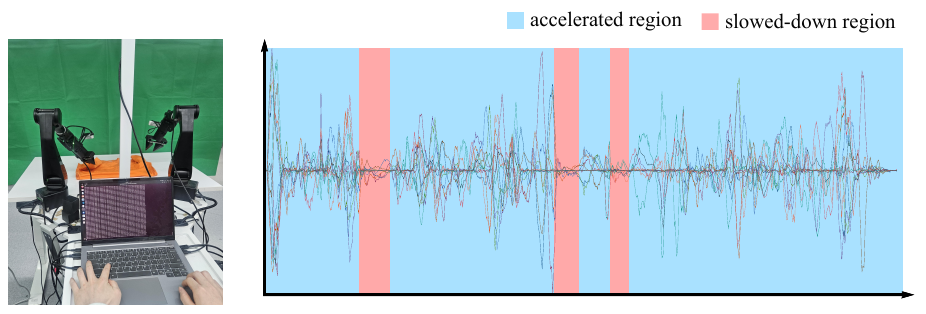}
    
    \caption{Human-in-the-loop speed modulation during closed-loop rollout. A human operator provides real-time ``throttle'' inputs to switch the robot between different execution speeds, slowing it down in precision-critical phases and accelerating it during low-risk motion. The plot shows joint-velocity traces from a real rollout; blue-shaded intervals indicate execution at $3\times$ speed, while red-shaded intervals indicate operator-triggered slowdowns to $1\times$. This procedure provides a practical way to collect speed-modulation supervision directly from real-world execution. \label{fig:human}}
\end{figure}

This method resorts to human intelligence. Ideally, we want an agent to be able to control the speed dynamically and learn to adjust when the current control policy leads to failure. We find humans to be good agents for this task. So in our setup, we give the operator of the robot a ``throttle'' input when the robot is running by a VLA. The operator will accelerate the real-time running speed when he/she feels comfortable on the current execution status, and slow-down the execution if he/she anticipates failure. We show the total time of the whole episode after each rollout, and encourage the operator to beat a daily optimization target.

To ease the control of speed adjustment and give our operator a chance to learn, the data collection is iterative. The throttle only adjusts the speed by at most a certain factor of speed-up relative to the current model's prediction. After one day's data collection, we train a regression model for speed adaptation and deploy it in the next day. Then the operator will face a higher base velocity and continue the optimization. When an episode encounters failure, we simply discard the labels in the surrounding region.

Because we are doing regression instead of density estimation from binary data, the model is harder to overfit, so we observed stable and predictable training of the speed modulation model. Also, the data can be used in an accumulative manner as the regression model will average out the different outcomes from different iterations. We setup an upper-bound on the absolute speed-up factor relative to demonstration (e.g. 3x or 4x) for safety considerations.

Theoretically, we can use RL algorithm instead of human operator in the process above. However, we currently do not find an algorithm with the same sample efficiency as real human. So we use this strategy as the current workaround and leave purely RL-based method to future work.
\section{Experiments}

We choose a set of tasks that are highly non-trivial while the current VLAs can solve. We list the details of the tasks in this section.

The first is \textbf{shirt-folding}, as shown below:
\begin{center}
    \includegraphics[width=0.8\linewidth]{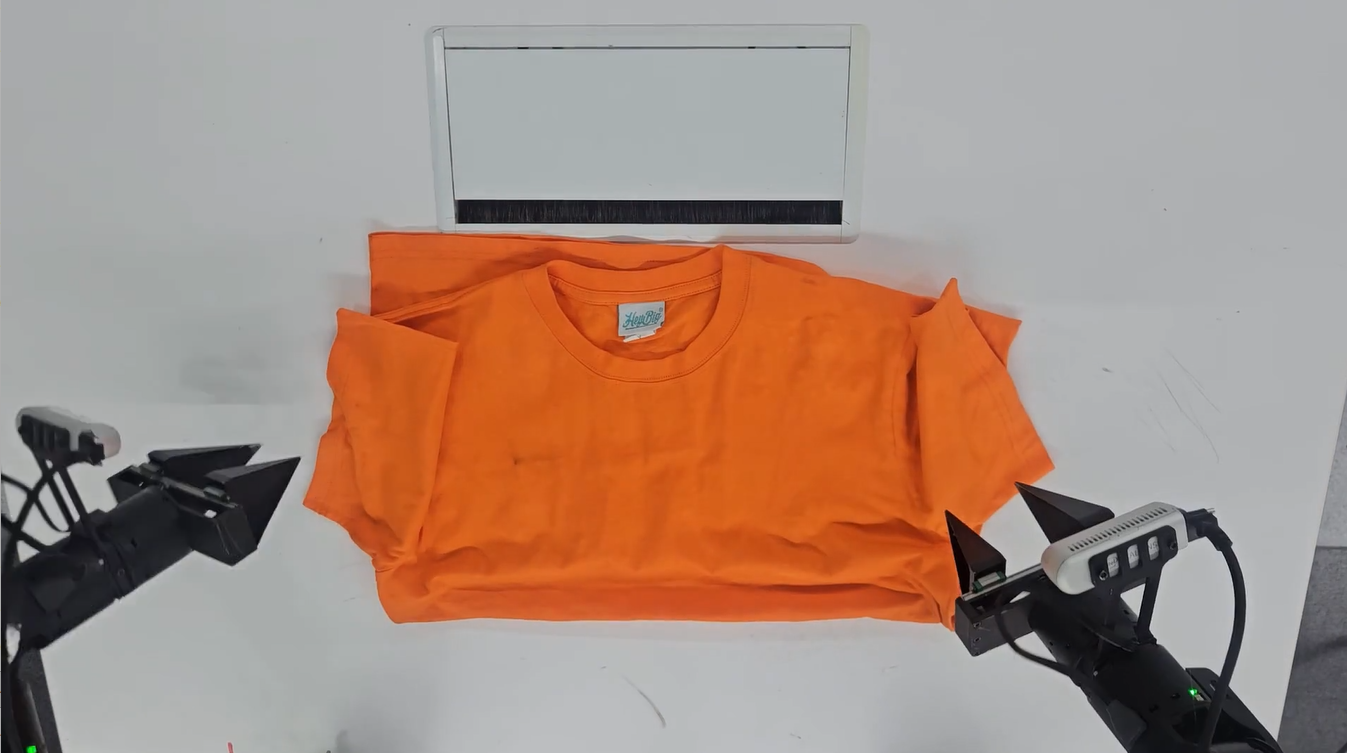}
\end{center}
In this task, the robot needs to fold the shirt into halves twice and put it to the upper right corner of the table. Before the first fold, it stretches the cloth on the table to flatten the wrinkles. It also needs to fold the sleeves before the second fold.

We choose this task as it is a classical benchmark in VLA learning and there are abundant demo videos from other methods that we can compare with on the Internet.

The second is \textbf{place-into-fixture}:
\begin{center}
    \includegraphics[width=0.8\linewidth]{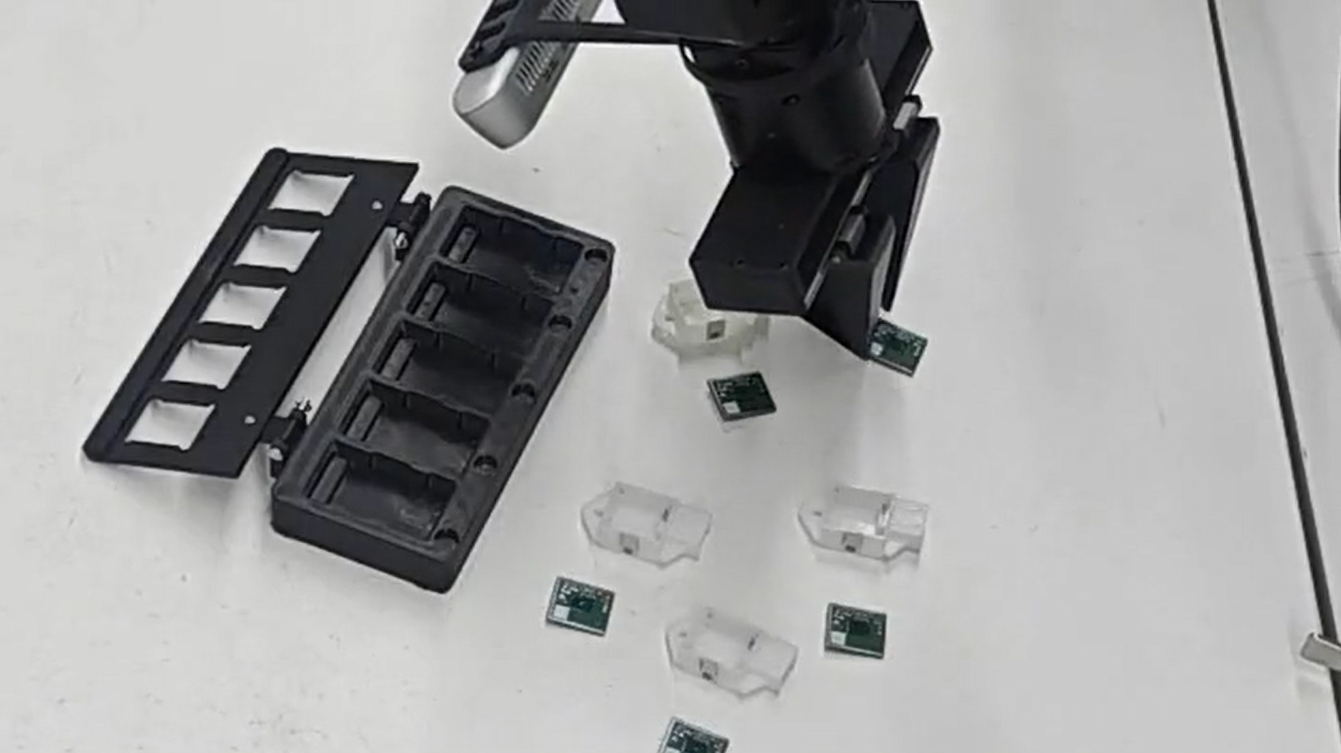}
\end{center}
In this task, the robot needs to pick the PCBs from the table into the white shields. Then the shields need to be placed into the fixture. There are totally 5 PCBs and shields, and the location of the parts can be random.

We choose this task as it requires a very high level of precision. Especially, when placing the PCB into the shield, three through-holes need to be precisely aligned to pins inside the shield. This is not easy even for human. Also, when placing the shield into the fixture, the margin is on the order of 0.2mm. The motion needs to be very precise to meet the alignment requirement.

In this task we expect the model to learn to slow down in the actions requiring contact and accuracy. And it really did.

The third task is \textbf{pick-and-latch}.
\begin{center}
    \includegraphics[width=0.8\linewidth]{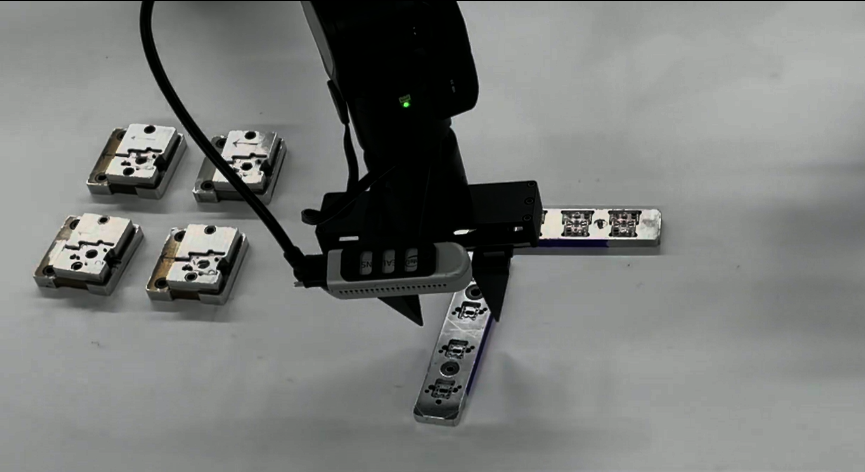}
\end{center}
In this task, the robot needs to pick four workpieces up from their individual fixtures, place them into a new fixture, and latch the fixture with another half on the top.

We choose this task as it touches the limit of our current VLA. Both the pick-and-place and the latching requires a great level of precision, making any attempt to accelerate the process non-trivial.

The outcome of our experiments are visualized in Fig.~\ref{fig:speedcompare}. We time the execution of a typical tele-operation based demonstration, a VLA model run, and the execution by a human. When the human does the task, we ask him/her to take the same action style as the robot for fairness. Also we instruct the human not to move too fast or too low, in a comfortable speed that can be repeated for a long time.

The figure shows near-human operating speed across all three tasks. Full videos are available at \url{https://dexmal.github.io/realtime-vla-v2/}, with RRD files tracing component operations during recording.

\section{Upperbound Analysis}

\begin{figure}
    \centering
    \includegraphics[width=1.0\linewidth]{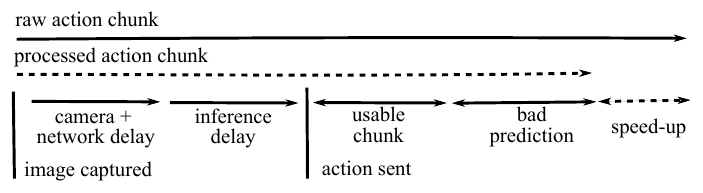}
    \caption{Relationship between delay time and max inference speed. \label{fig:delaybound}
    }
\end{figure}

Another issue we want to discuss is whether we can setup an upperbound on how fast we can run the VLA models on real robots. We propose a framework of argument, which is reminiscent to the roofline model in computation, that decomposes the motion in different phases with corresponding bounding factors.

The \textbf{motion bounded} area indicates the acceleration, velocity, or jerk is approaching the hardware's limit in endurable execution. They are the already optimal regions that we cannot expect further speedup.

The \textbf{control bounded} area indicates the region where the hardware can run faster but the latency of control is so high that success rate cannot be maintained. To better describe this mechanism, we draw a control loop diagram in Fig.~\ref{fig:delaybound}. The model generates an action chunk given a image input. The prefix is buried by the receding horizon. Then there is a segment of the chunk that is usable to control the robot. Afterwards, in the late part of the segment, the model's prediction becomes unreliable so that we cannot use it. Finally, when we speed up the chunk, the effective length of it will shrink accordingly. It can be seen that the system is effective only when the length of the usable chunk is positive. When the part of motion is so intricate that the ``bad prediction'' part increases, we have to reduce the speed-up factor to accommodate for a shorter overhead prefix. This is the part that we will call ``control bounded''.

In one robot run, we can segment the total trajectory into segments that are either motion bounded or control bounded. If this is achieved, we will feel confident that the speed adaptation algorithm has already done its job. Then if we want to further increase throughput, we either (1) improve the hardware so that the motion bounded regions move faster (2) improve the model so that it tolerates longer control feedback delay. This gives us a principled framework for studying execution acceleration techniques.

\begin{figure}
    \centering
    \includegraphics[width=1.0\linewidth]{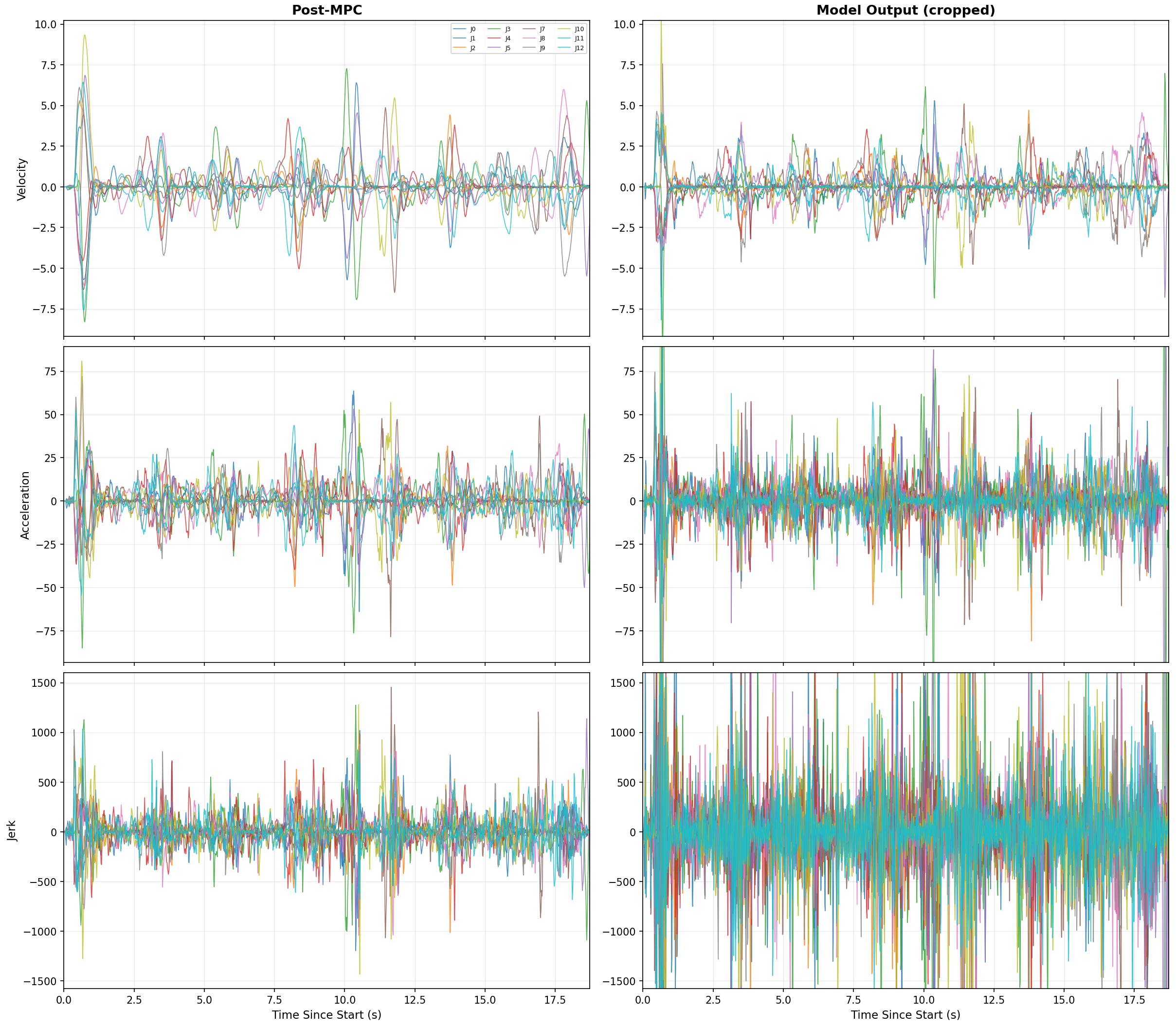}
    \caption{Velocity, acceleration and jerk after and before post-processing.\label{fig:trajcompare}}
\end{figure}

To get an understanding on how our current method approaches these bounds, we plot the velocity, acceleration, and jerk profile of a fold-the-shirt run in Fig.~\ref{fig:trajcompare}. On the right side is the profile of the model's output if we directly squeeze it into the target total time. On the left is the profile of our post-processed trajectory. We can see that the scale of acceleration and jerk is reduced and evenly distributed throughout the run. According to the readings and our experience with the robot, in most parts it is already motion-bound. This is a clear signal for optimality at the current stage.

\section{Conclusion}

In this paper, we demonstrate how to speed up the execution of VLA policies on robots to achieve near-human performance. Our techniques include both system-level and data-driven approaches. With all our methods being implemented, we achieve impressive results in realistic robot tasks. We hope our work will inspire future researchers in this direction.

{
    \small
    \bibliographystyle{ieeenat_fullname}
    \bibliography{main}
}


\end{document}